# Automatic Image Pixel Clustering based on Mussels Wandering Optimization

Xin Zhong[1], Frank Y. Shih[2], Xiwang Guo[3]
1. Department of Computer Science, University of Nebraska at Omaha, Omaha NE 68182
2. Department of Computer Science, New Jersey Institute of Technology, Newark, NJ 07102
3. Department of Electrical and Computer Engineering, New Jersey Institute of Technology, Newark, NJ 07102

Contact: xzhong@unomaha.edu

**Abstract** Image segmentation as a clustering problem is to identify pixel groups on an image without any preliminary labels available. It remains a challenge in machine vision because of the variations in size and shape of image segments. Furthermore, determining the segment number in an image is NP-hard without prior knowledge of the image content. This paper presents an automatic color image pixel clustering scheme based on mussels wandering optimization. By applying an activation variable to determine the number of clusters along with the cluster centers optimization, an image is segmented with minimal prior knowledge and human intervention. By revising the within- and between-class sum of squares ratio for random natural image contents, we provide a novel fitness function for image pixel clustering tasks. Comprehensive empirical studies of the proposed scheme against other state-of-the-art competitors on synthetic data and the ASD dataset have demonstrated the promising performance of the proposed scheme.

# 1 Introduction

Image segmentation aims to group image pixels into disjoint regions in such a way that the pixels in the same region are more similar to each other than to those in other regions. Fig. 1 shows two images with their segmented results. The similarity between pixels can be defined by various image features, such as luminance, color, and spatial position. Image segmentation simulates a nature in human vision that a visual field is organized into patterns containing objects and surroundings. Hence, it supports different core applications in computer vision, including scene understanding, content-based image retrieval, and image classification.

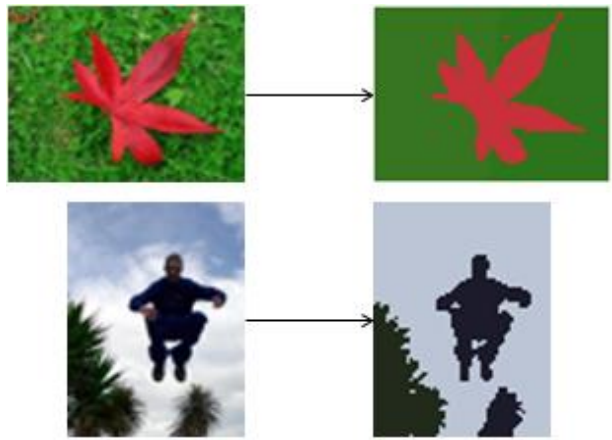

Fig. 1. Natural images and their segmented results.

On the other hand, clustering analysis is the process of indentifying natural groups in an unlabeled sample set of multidimensional data. From this point of view, image segmentation can be performed as a clustering problem, where the sample set contains image pixels, and the dimensionalities describing the samples are image features [9]. The identified groups or clusters are referred to as the patterns or segments of images.

Clustering algorithms can be categorized into being hierarchical and partitional based on the solving process. The hierarchical clustering methods intend to solve the clustering problem via organizing a tree structure (i.e. the dendrogram), where each node represents a cluster to various levels of granularity [7]. As a bottom-up process, hierarchical clustering algorithms keep merging nodes of the tree until a cut-off threshold has been reached. In contrast, partitional clustering methods decompose the dataset with a given partition number. In a top-down manner, partitional clustering algorithms search for the best decomposition scheme of the dataset based on some given objective functions.

Clustering algorithms can also be divided into crisp solutions and fuzzy solutions based on the segmented results [5]. Crisp clustering produces nonoverlapping clusters, in which a sample point can only be clustered into a single cluster. Fuzzy clustering, in contrast, a sample point may belong to multiple clusters with a fuzzy membership function [9]. In this paper we focus on crisp clustering in image segmentation.

Pixel clustering remains a challenging task in machine vision because of the variations in size and shape of image segments. Determining the optimal number of clusters in most of the prevailing algorithms has been proved to be NP-hard, and often requires prior knowledge of the dataset [10]. It is extremely difficult to summarize a prior knowledge for the optimal number of segments due to the randomness in natural image contents. Furthermore, traditional clustering algorithms, such as K-means, suffer from drawbacks, including the dependence upon the initial selection of centers, dead unit, and local minima convergence [21]. To address such issues, evolutionary clustering algorithms have been applied during recent years. By viewing the clustering problem as an optimization process, the evolutionary algorithms deliver promising performance due to the capability of global optimization, as well as the effective mechanism derived from the biological system.

Early work on automatic clustering relies on conventional genetic algorithm [19, 20]. However, the outcome is unstable and unsatisfactory when it comes to image segmentation. Lai and Chang [14] proposed a hierarchical evolutionary based algorithm for medical image segmentation. By using a variant of conventional genetic algorithm, it produces medical image segments automatically. Omran *et al.* [16] proposed an image segmentation scheme based on particle swarm optimization (PSO) [11]. The scheme optimizes the centers of a pre-defined number of clusters by applying the intelligence of swarm particles. Omran *et al.* [18] improved the PSO-based image segmentation scheme by allowing automatic determination of the cluster number. Starting with a large number of clusters, the algorithm refines the centers and determines the cluster numbers simultaneously with a binary PSO. Besides PSO, differential evolution (DE) [19] was also applied in this field. Omran *et al.* [17] proposed a nonautomatic clustering scheme based on DE for image classification. Das *et al.* [5] developed an evolutionary fuzzy pixel clustering scheme for grayscale images. A variant of DE has been used to determine the number of segments in an image as well as to refine the cluster centers. Moreover, mussels wandering optimization (MWO) [2] was proposed with promising performance for clustering problems. Kang *et al*. [12] developed a nonautomatic clustering method based on MWO by applying two objective functions. In comparisons, MWO is validated to have better performance than PSO when dealing with the cluster center refinement.

In this paper, we present an automatic color image pixel clustering scheme based on MWO. First, we show that under the mussels' representation, MWO can be used for color image segmentation. By applying an activation variable to determine the number of clusters along with the cluster centers optimization, an image is segmented with minimal prior knowledge and human intervention. Second, by revising the within- and between-class sum of squares to a balanced form to deal with random natural image contents, we propose a novel fitness function that indicates the ratio of within-class sum of squares over between-class sum of squares. Third, we provide comprehensive empirical studies of both the proposed method and some state-of-the-art competitors on synthetic data and a natural large-scale image dataset.

The remainder of this paper is organized as follows. Section 2 describes the proposed automatic color image pixel clustering in details including the groundwork and the algorithm. Section 3 presents the experimental simulation results on both synthetic and natural images. Conclusions are drawn in Section 4.

## 2 The proposed MWO-based automatic pixel clustering algorithm

**2.1 Mussels wandering optimization (MWO)**

Inspired by the leisure movement that forms pattern in mussels' habitat, MWO is proposed as an optimization scheme based on swarm intelligence. The literature in [2, 12] has shown its promising performance developed from the effective mechanisms and the desired characteristics of biological evolutions. In MWO, a landscape-level evolutionary mechanism of the mussels' distribution pattern is simulated through a stochastic decision and a levy walk. MWO has been applied in comprehensive problems like global optimization, hence applying it in automatic pixel clustering schemes can avoid the local minima convergence. In addition, some studies on human vision have pointed out that photons that strike the retina would form a geometrical landscape or surface [15]. Some methods in computer vision capture this neural landscape information by the second order of image gradients [8]. In this paper, MWO is adopted for its landscape-level evolutions, and therefore provides a well-defined description of neural images in pixel clustering.

MWO can be divided into several steps. The first step is to initialize a population of mussels with an evaluation for each based on a fitness function. Mussels with a good fitness value are used to update other mussels in the next iteration, so the overall evolution is guided toward a better direction. A levy walk, between 0 and 1, is defined to control each mussel's displacement. Remarkably, any update should not exceed the search range, thus avoiding the mussels going to an undesirable field.

**2.2 Feature selection and distance measure**

Selecting the dimensionalities to describe each pixel is crucial in the pixel clustering task. For a color image in digital format, only limited information can be derived. Hence, we select low-level perceptual set in human vision to simulate neural perceptual nature. In psychology, the low-level visual perceptual set includes proximity, color, and contrast. First, we use *XY* location in Cartesian coordinate system to capture the spatial position, so that close *XY* locations can ensure spatial proximity when it comes to a clustering task. Using *XY* coordinates for spatial information was applied in the nonautomatic pixel clustering method in [3] for simplicity and effectiveness.

As to include image color information, one can choose values in various digital image channels, for example, *RGB* image channels or *CieLab* color channels. The *RGB* follows the color receptor in human eye, and the bit-depth determines the granularity of color. The *CieLab* describes all perceivable colors in luminance *L*, and *a* and *b* for the color components green-red and blue-yellow, respectively. *CieLab* color has been considered as the closest simulation to human color perception. There are additional options to be considered, such as *YUV* and *HSV* color space. In the clustering, close color values imply similar appearance of pixels. In this paper, we apply *RGBXY* space to describe each pixel by combining both color and position information.

Another factor in the low-level visual perceptual set is the contrast, which has been considered as the most important factor in an image to convey information. To measure the contrast between pixels is to compute a distance between them. In image context, several distance measurements have been studied for different purposes [22]. For example, a chessboard distance to include a local 8-connectivity information, and a max-arc or a minimum barrier distance to describe the graph path based information. In this paper, we use the *RGBXY* or *LABXY* feature space for pixels, and Euclidean distance measure for its simulation to human perception.

**2.3 Mussels' representation**

In the proposed method, a mussel consists of four major properties related to a clustering solution: a position consists of $K_{max}$ of a six-dimensional value representing cluster centers and their activations, where $K_{max}$ is a user-defined maximum number of segments, a fitness value, a levy walk value, and the output including the segmentation result, activated centers, and the optimal number of clusters.

A single mussel using *RGBXY* in the feature space is presented in Table 1. Besides $(RGBXY)_{kc}$, where $kc = 1, 2, 3, \ldots, K_{max}$ for pixel centers, a mussel's position contains an additional activation dimension $A_{kc}$, a real number between 0 and 1 to determine if its corresponding center will be used during the clustering. The activation threshold is set to be 0.5. For example, if $K_{max}$ is 9 and $A_1$, $A_3$ and $A_5$ are larger than 0.5, only three centers 1, 3 and 5 will be used among the total of nine centers to generate three clusters. Using the activation value along with the cluster centers in the mussels' position not only can refine the clustering result, but also can determine a good segment number.

For each dimension during the update, values exceeding their ranges will be forced to its minimum of maximum. For example, any $R_{kc}$ larger than 255 will be set to 255, and any $A_{kc}$ smaller than 0 will be set to zero. Furthermore, if less than two $A_{kc}$ values are smaller than the threshold for a mussel, current largest two $A_{kc}$ values will be set to a random number from 0.5 to 1 during the update, thus at least two clusters will be produced for any segmentation. Then, a mussel contains a fitness function $f$ with its position, and a mussel's levy walk $L$ is computed according to [2, 12], which will be presented in the next section.

Table 1. A single mussel's properties

| | |
|---|---|
| position | $\begin{bmatrix} R_1 & G_1 & B_1 & X_1 & Y_1 & A_1 \\ R_2 & G_2 & B_2 & X_2 & Y_2 & A_2 \\ ... & ... & ... & ... & ... & ... \\ R_{k_{max}} & G_{k_{max}} & B_{k_{max}} & X_{k_{max}} & Y_{k_{max}} & A_{k_{max}} \end{bmatrix}$ |
| fitness | $f = \text{fitness(position)}$ |
| levy walk | $L = \gamma(1 - rand)^{-1/(\mu-1)}$ |
| output | segmentation, centers, number of clusters |

## 2.4 The Fitness Function

Classic K-means and other evolutionary clustering methods [12] aiming at center refinement intend to minimize the within-class sum of squares *W* as:

$$W = \sum_{i=1}^{k} \sum_{j=1}^{n_i} (x_{i,j} - \bar{x}_i)(x_{i,j} - \bar{x}_i)^T \tag{1}$$

where $x_{i,j}$ is the *j*-th data point from the *i*-th cluster with $n_i$ data points, $\bar{x}_i$ denotes the center for the *i*-th cluster, and *k* is the cluster number. Minimizing *W* can secure that the similarity of members in the same cluster will be maximized. However, as the within-class sum of squares is not convex, using it as the fitness function can possibly lead the clustering result to multiple local minima [13, 21].

Since clustering aims to not only maximize the similarity inside each cluster, but also minimize the similarity between clusters, we consider our fitness function using both the within- and between-class sum of squares. Traditional between-class sum of squares *B* is defined as:

$$B = \sum_{i=1}^{k} n_i (\bar{x}_i - \bar{x})(\bar{x}_i - \bar{x})^T \tag{2}$$

where $\bar{x}$ is the mean of all the points. The $W$ and $B$ are two decompositions of the total sum of squares $T$, and the summation of $W$ and $B$ is $T$.

$$T = \sum_{i=1}^{k} \sum_{j=1}^{n_i} (x_{i,j} - \bar{x})(x_{i,j} - \bar{x})^T \qquad (3)$$

Our goal is to minimize $W$ and simultaneously maximize $B$. Therefore, the ratio of $W/B$ is targeted to be minimized [6]. However, as image pixel groups may vary in size, $B$ can be dominated by a cluster with a large $n_i$, in which the clustering will be looking for groups separating the large cluster from the others, but will not evaluate much about how well other relatively small clusters are separated. To balance the effect of each cluster regardless of its size, we revise the between-class sum of square $B'$ as

$$B' = \bar{n} \sum_{i=1}^{k} (\bar{x}_i - \bar{x})(\bar{x}_i - \bar{x})^T \qquad (4)$$

where $\bar{n}$ is the mean of points per cluster. However, using $B'$ does not satisfy the decomposition of $T$ into $W$ and $B'$. Hence, $W$ is revised as $W'$ to restore the identity.

$$W' = \bar{n} \sum_{i=1}^{k} \frac{1}{n_i} \sum_{j=1}^{n_i} (x_{i,j} - \bar{x}_i)(x_{i,j} - \bar{x}_i)^T \qquad (5)$$

Under the definitions of $B'$ and $W'$, the new total sum of squares $T'$ is

$$T' = W' + B' = \bar{n} \sum_{i=1}^{k} \frac{1}{n_i} \sum_{j=1}^{n_i} (x_{i,j} - \bar{x})(x_{i,j} - \bar{x})^T \qquad (6)$$

where $T'$, $B'$ and $W'$ are all invariants regardless the cluster sizes.

The fitness value $RF$ is the ratio of $W'$ over $B'$. Our objective is to minimize $RF$, which is to maximize the similarity inside a cluster and minimize the similarity between clusters at the same time.

$$RF = \frac{W'}{B'} \qquad (7)$$

**2.5 The proposed automatic pixel clustering based on MWO**

The proposed automatic pixel clustering algorithm based on MWO is summarized as follows:

1. Initialize a population of $N$ mussels, where there are $K_{max}$ centers randomized within the ranges. Compute the levy walk for each mussel.

2. For every mussel, use the activated positions as the centers to cluster the pixels (i.e., assign each pixel to its closest center.) and compute the fitness value via Eq. (7).

3. Sort all the mussels based on their fitness values, find the top $t$ mussels, and compute their mean position $P_t$.

4. Update each of the positions $P_i$ for the remaining nontop mussels by the difference between $P_i$ and $P_t$ as well as its levy walk $L$.

$$P_i' = P_i + L(P_t - P_i) \tag{8}$$

5. Stop the program and output the sorted mussels if a terminate condition is reached. Otherwise, go to step 2.

## 3 Experimental simulation

In this section we present simulation results on different types of data by applying the proposed automatic pixel clustering method (denoted ACMWO) and other evolutionary clustering methods, including a GA-based automatic pixel clustering method [14] (denoted ACGA), a PSO-based automatic pixel clustering method [18] (denoted ACPSO), and a DE-based automatic pixel clustering method [5] (denoted ACDE). Note that in ACMWO, we set $\gamma = 1$ and $\mu = 2$ in the mussel's levy walk computation. The maximum number of clusters $K_{max}$ is set to 15.

Most of the measurement indexes for clustering results are optimized, and the maximum or minimum values of these indexes indicate the appropriate partitions. We adopt the DB index in [6] to evaluate the clustering result for its prevalence. The DB index essentially indicates the ratio of the sum of within-cluster scatter to between-cluster separation, in which the within $i$-th cluster scatter $S_{i,q}$ and the between $i$-th and $j$-th cluster distance $d_{i,j,t}$ are respectively defined as:

$$S_{i,q} = (\frac{1}{n_i}\sum_{x \in C_i} ||x - \bar{x}_i||_2^q)^{1/q} \qquad (9)$$

$$d_{i,j,t} = ||\bar{x}_i - \bar{x}_j||_t \qquad (10)$$

where $\bar{x}$ is the cluster center, $n_i$ is the number of points in $i$-th cluster $C_i$, and $q, t \geq 1$ are the selected orders. The ratio $R_{i,q,t}$ of $i$-th within-cluster scatter over between $i$-th and $j$-th cluster distances is defined as:

$$R_{i,q,t} = max_{j \neq i}(\frac{S_{i,q} + S_{j,q}}{d_{i,j,t}}) \qquad (11)$$

The DB index in terms of a cluster number $k$ is defined as:

$$\text{DB} = \frac{1}{k}\sum_{i=1}^{k} R_{i,q,t} \qquad (12)$$

Note that a smaller DB index indicates a better clustering result.

### 3.1 Simulations on synthetic data

We first present the simulation results on a generated two-dimensional sample data for a better visualization. The sample data is produced with three denser areas along with sparser points in between as shown in Fig. 2.

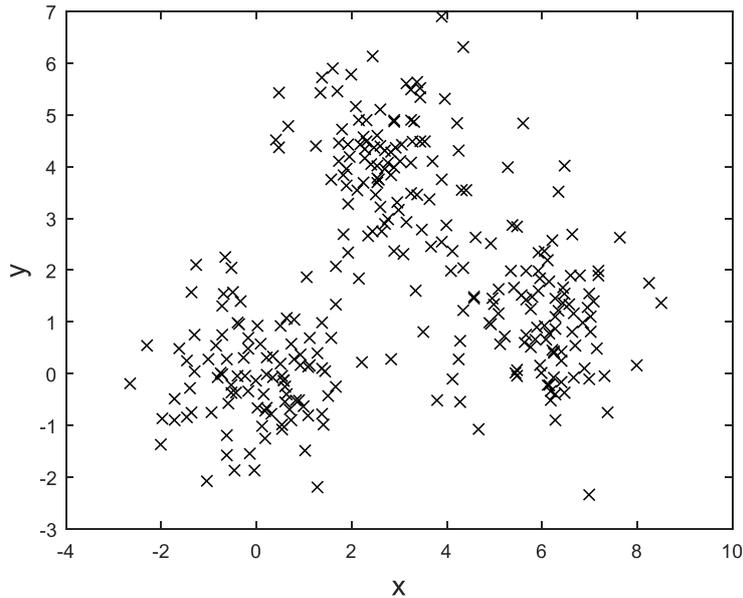

Fig. 2. The two-dimensional sample data.

Our proposed ACMWO method is compared against ACGA [14], ACPSO [18], and ACDE [5]. Each algorithm is applied 20 times to compute the average result. Fig. 3 visually compares the clustering for each algorithm, where the cluster centers are depicted as the black circle "o". Note that ACGA and ACPSO both produce two clusters, while ACDE and ACMWO successfully identify the three clutter areas in the generated sample. Visually we can observe that ACMWO locates the centers (i.e. the density) more accurately than the competitors. To quantitatively evaluate the centers refinement for the algorithms, their average DB indices of the final clustering results are listed in Table 2, where we conclude that our proposed ACMWO presents the best clustering result with the smallest DB index value. In addition, we plot the DB index of its global best solution for each algorithm over 200 iterations to visualize the evolution rate. The ACGA and ACPSO stabilize fast before 20 iterations, and the ACDE and ACMWO stabilize over 60 iterations.

Table 2. The average DB for the sample data

| Algorithm  | ACGA   | ACPSO  | ACDE   | ACMWO      |
|------------|--------|--------|--------|------------|
| Average DB | 0.6307 | 0.6309 | 0.5591 | **0.5288** |

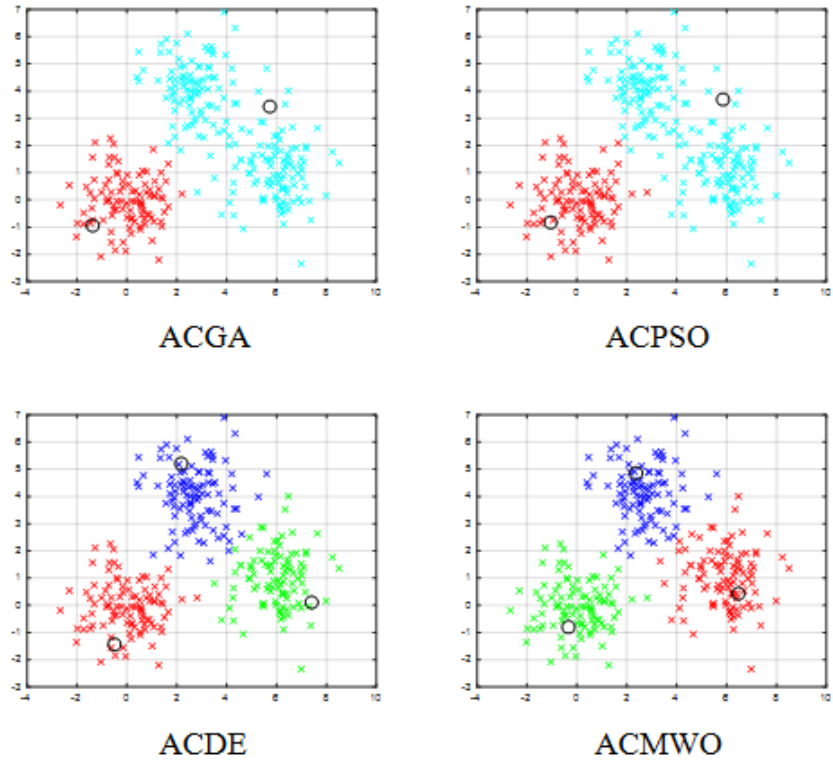

Fig. 3. Visual comparison of the clustering results.

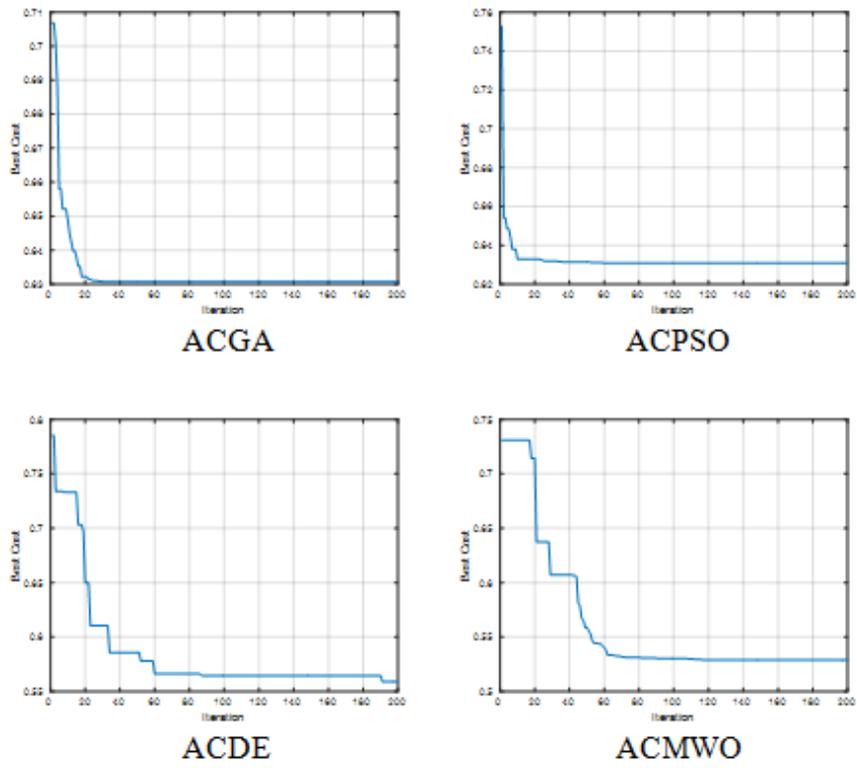

Fig. 4. DB indices of the best solution on the sample data over 200 iterations.

**3.2 Reponses to colors**

Next, we present the simulation results on a synthetic image as shown in Fig. 5 to further illustrate the efficiency of ACMWO. The synthetic image is created as three squares with color red, green and blue (*RGB*) on a black background. Performance on this image indicates the responses of an algorithm towards basic *RGB* color. An ideal pixel clustering should produce a DB index of zero in this situation if it is correctly clustered with four clusters (i.e. black, red, green, and blue). Because the within class sum is zero for pure colors.

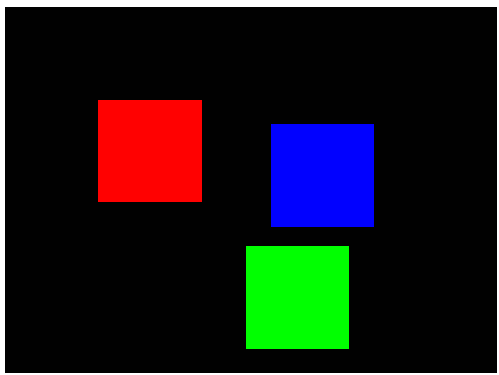

Fig. 5. A synthetic image with RGB color.

The clustering results of different algorithms are presented in Fig. 6. We display the segments by labeling a grayscale for each cluster, so that the appearance of pixel members within a cluster is identical. It is observed that only the proposed AMWO perfectly identifies the four segments, while the ACGA produces two clusters and segments the green square, the ACPSO produces two clusters and segments the red square, and the ACDE produces three clusters and segments the red and blue squares.

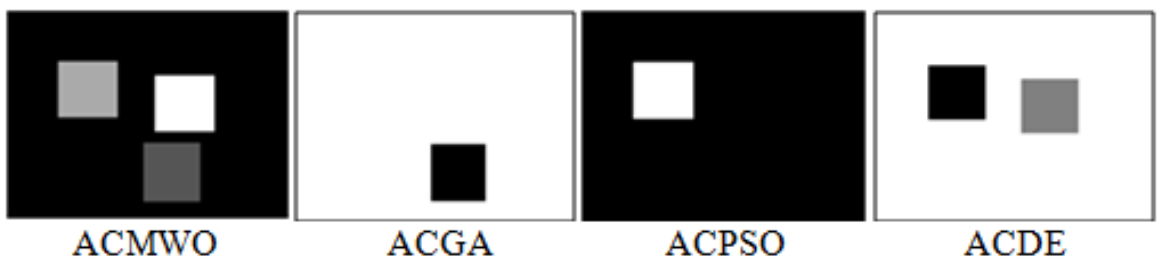

Fig. 6. The clustering results on the synthetic image.

Quantitatively, the average DB indices of the final clustering results for ACMWO, ACGA, ACPSO and ACDE are 0, 0.3408, 0.3408, and 0.2532, respectively, from which we conclude that ACMWO outperforms the competitors in terms of the responses towards basic *RGB* color for pixel clustering. Again, we plot the DB indices of the algorithms' global best solution over 200 iterations, where we can visualize that ACMWO stabilizes fast to 0.

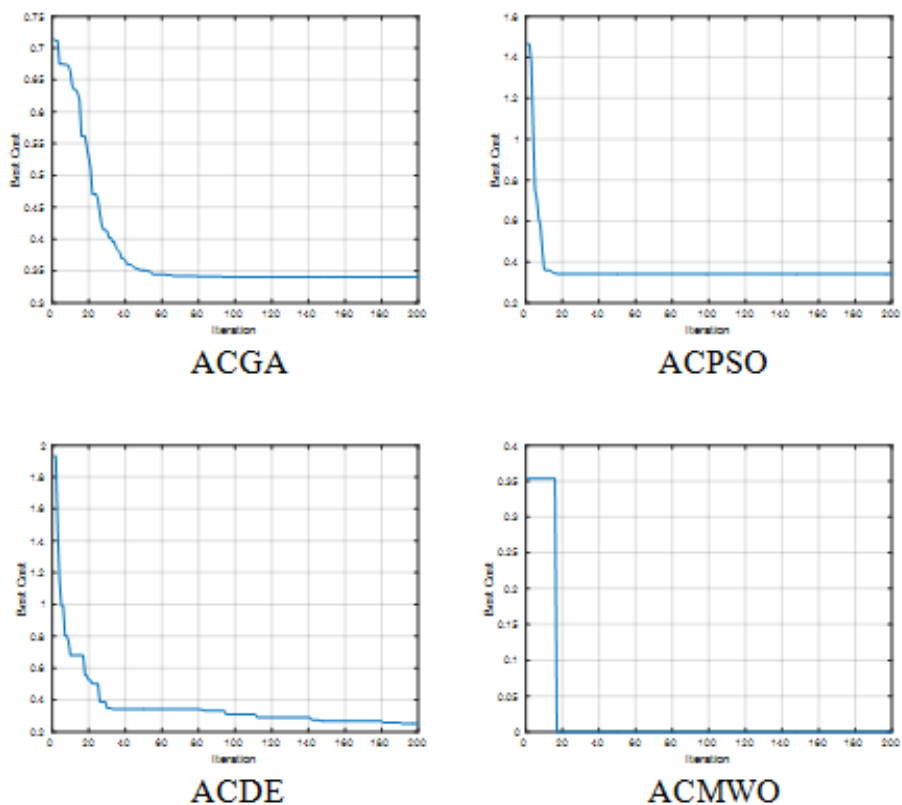

Fig. 7. DB indices of the best solution on the synthetic image over 200 iterations.

Another example of containing various colors is shown in Fig. 8, where besides *RGB*, we create orange, yellow, and purple on a white background. The correct segment number in this case is seven. All other colors can be created with *RGB*; for example, purple consists of red and blue components. We add these comprehensive colors to bring some challenges to the testing algorithms.

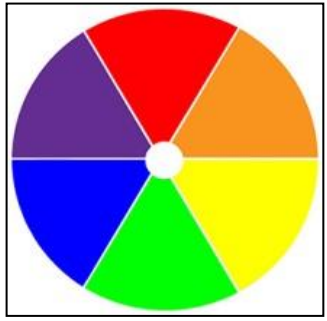

Fig. 8. A synthetic image of six colors in the white background.

The clustering results of different algorithms are presented in Fig. 9. The ACGA only segments white from other colors and produces a DB index of 0.7803. The ACPSO only segments blue and purple from other colors and produces a DB index of 0.7375. The ACDE segments green as a cluster, red, orange and yellow (the colors with red components) as a cluster, blue and purple as a cluster, and the background white as the last cluster, and produces a DB index of 0.58954. The ACMWO creates five segments with white, blue/purple, red, orange/yellow and green color and produces a DB index of 0.3999. Therefore, the ACMWO outperforms the competitors with comprehensive colors.

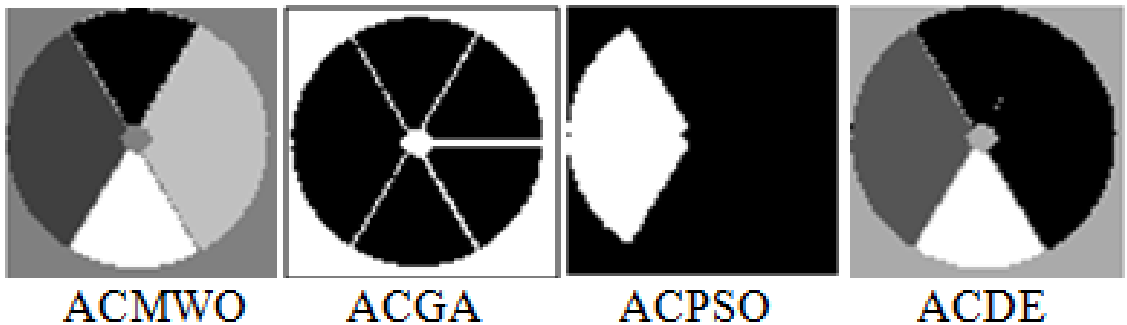

Fig. 9. The clustering results of applying different algorithms.

### 3.3 Experiments on natural images

We run ACMWO, ACPSO and ACDE on the ASD [1] image dataset, which contains 1,000 natural color images with various image sizes and image contents, as providing a challenge to each algorithm with unknown number of segments and different image appearance. To empirically evaluate the performance on this large-scale data, we cluster the pixels for each image and compute the mean DB index as well as the variance, which not only provides the average performance, but also shows the stability for each algorithm. The quantitative results are listed in Table 3, from which we observe that ACMWO shows the lowest mean DB index and the smallest variance. It indicates that ACMWO outperforms the competitors in automatic image pixel clustering tasks with both a better average performance and a more stable segmentation.

Table 3. Performance of ACMWO against the competitors.

| Method / Evaluatio | ACMWO | ACPSO | ACDE |
|---|---|---|---|
| Mean | **0.6895** | 0.7039 | 0.7196 |
| Variance | **0.0066** | 0.0083 | 0.0097 |

Fig. 10 shows some images from the ASD dataset, along with the pixel clustering results of ACMWO, ACDE and ACPSO for visual comparisons. Each image segment is displayed by its mean color for better visualization. It is observed that the segments of ACMWO are closer to the divisions in human vision system. In other words, the clustering results of ACMWO produce better segmentation than ACDE and ACPSO, as much alike appearance inside each segment and more distinct appearance between segments.

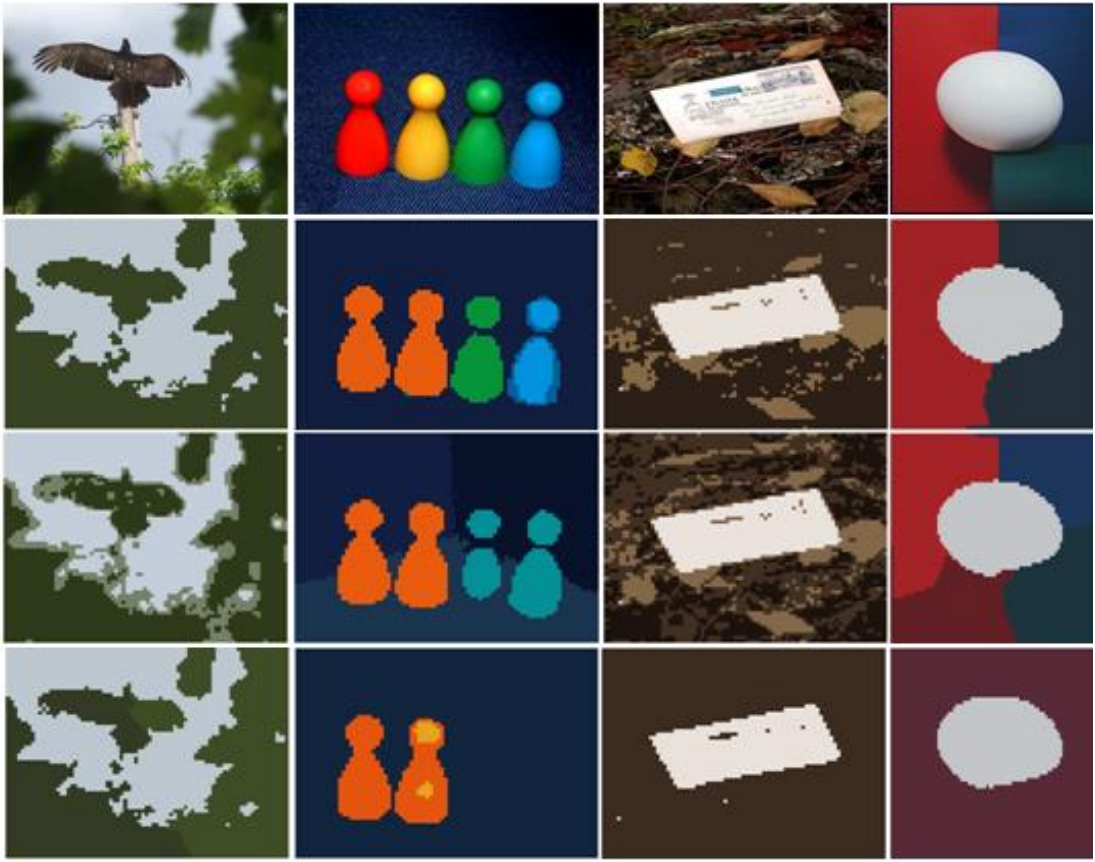

Fig. 10. Some images from the ASD dataset and their segmentation results. Row 1: The original image. Row 2: Segmentation of ACMWO. Row 3: Segmentation of ACDE. Row 4: Segmentation of ACPSO.

## 4 Conclusions

This paper presents an automatic color image pixel clustering scheme based on MWO. We demonstrate the supreme of MWO for color image segmentation by the mussels' representation with activations. A novel fitness function, which balances the effect of different clusters for random natural images, is both analytically and empirically confirmed. Experimental results present a promising performance of the proposed scheme. Future research will focus on the distance measure for color images. Applying some newly developed image distance function in the evolution, such as geodesic distance and minimum barrier distance, is still an open question. The difficulty is to determine or develop novel clustering evaluation indices with different distance function.

**References**

1. Achanta R, Hemami S, Estrada F, Susstrunk S (2009) Frequency-tuned salient region detection. Proc. IEEE Conference on Computer Vision and Pattern Recognition, In pp. 1597-1604.

2. An J, Kang Q, Wang L, Wu QD (2013) Mussels wandering optimization: an ecologically inspired algorithm for global optimization. Cogn. Comput. 5(2): 188–199.

3. Achanta R, Shaji A, Smith K, Lucchi A, Fua P, Süsstrunk S (2012) SLIC superpixels compared to state-of-the-art superpixel methods. IEEE trans. pattern analysis and machine intelligence 34(11): 2274-2282.

4. Bandyopadhyay S, Maulik U (2002) Genetic clustering for automatic evolution of clusters and application to image classification. Pattern Recognit. 35: 1197–1208.

5. Das S, Abraham A, Konar A (2008) Automatic clustering using an improved differential evolution algorithm. IEEE Trans. systems, man, and cybernetics-Part A: Systems and Humans, 38(1):.218-237.

6. Davies DL, Bouldin DW (1979) A cluster separation measure. IEEE Trans. pattern analysis and machine intelligence 1(2): 224–227.

7. Guo G, Chen S, Chen L (2012) Soft subspace clustering with an improved feature weight self-adjustment mechanism. Int. Machine Learning and Cybernetics 3(1): 39-49.

8. Huang D, Zhu C, Wang Y, Chen L (2014) HSOG: a novel local image descriptor based on histograms of the second-order gradients. IEEE Trans. Image Processing 23(11): 4680-4695.

9. Jain AK, Murty MN, Flynn PJ (1999) Data clustering, a review. ACM Computer Survey 31(3): 264–323.

10. Jain AK, Duin R, Mao J (2000) Statistical Pattern Recognition: A Review. IEEE Trans. Pattern Analysis and Machine Intelligence 22(1): 4-37.

11. Kennedy J (2010) Particle swarm optimization. Encyclopedia of Machine Learning, Springer US: 760-766.

12. Kang Q, Liu S, Zhou M, Li S (2016) A weight-incorporated similarity-based clustering ensemble method based on swarm intelligence. Knowledge-Based Systems 104: 156-164.

13. Kao Y, Zahara E, Kao I (2007) A Hybridized Approach to Data Clustering Expert Syst Appl 34: 1754–1762.

14. Lai CC, Chang CY (2009) A hierarchical evolutionary algorithm for automatic medical image segmentation. Expert Systems with Applications 36(1): 248-259.

15. Morgan MJ (2011) Features and the 'primal sketch'. Vis. Res. 51(7): 738–753.

16. Omran M, Engelbrecht AP, Salman A (2005) Particle swarm optimization method for image clustering. Int. Pattern Recognition and Artificial Intelligence, 19(3): 297-321.

17. Omran M, Engelbrecht AP, Salman A (2005) Differential evolution methods for unsupervised image classification. Proc. Evolutionary Computation, In 2: 966-973.

18. Omran M, Salman A, Engelbrecht AP (2006) Dynamic clustering using particle swarm optimization with application in image segmentation. Pattern Analysis and Applications 8(4): pp. 332.

19. Price K, Storn R, Lampinen J (2005) Differential Evolution – A Practical Approach to Global Optimization. Springer, Berlin.

20. Sarkar M, Yegnanarayana B, Khemani D (1997) A clustering algorithm using an evolutionary programming-based approach. Pattern Recognit. Lett. 18: 975–986.

21. Tsai CY, Chiu CC (2008) Developing a feature weight self-adjustment mechanism for a K-means clustering algorithm. Computational statistics & data analysis 52(10): 4658-4672.
22. Zhang J, Sclaroff S (2016) Exploiting surroundedness for saliency detection: a Boolean map approach. IEEE trans. pattern analysis and machine intelligence, 38(5): 889-902.